\documentclass[%
reprint,
 amsmath,amssymb,
 aps,
showkeys,
]{revtex4-2}
\usepackage{siunitx}
\usepackage{graphicx}
\usepackage{dcolumn}
\usepackage{bm}
\usepackage[T1]{fontenc}

\usepackage[utf8]{inputenc}
\usepackage{soul}
\usepackage{xcolor}
\sethlcolor{yellow}

\begin{document}

\preprint{APS/123-QED}

\title{Estimation of Energy-dissipation Lower-bounds for Neuromorphic Learning-in-memory}

\author{Zihao Chen}
\affiliation{%
 Electrical \& Systems Engineering, Washington University in Saint Louis, St. Louis, MO 63130
}%
\author{Faiek Ahsan}
\affiliation{%
 Electrical \& Systems Engineering, Washington University in Saint Louis, St. Louis, MO 63130
}%

\author{Johannes Leugering}
\affiliation{%
 Bioengineering, University of California San Diego, La Jolla, CA 92093
}%
\author{Gert Cauwenberghs}
\affiliation{%
 Bioengineering, University of California San Diego, La Jolla, CA 92093
}%
\author{Shantanu Chakrabartty}
\email{All correspondence regarding this manuscript should be addressed to shantanu@wustl.edu.}
\affiliation{%
 Electrical \& Systems Engineering, Washington University in Saint Louis, St. Louis, MO 63130
}%

\date{\today}

\begin{abstract}
Neuromorphic or neurally-inspired optimizers rely on local but parallel parameter updates to solve problems that range from quadratic programming to Ising machines. An ideal realization of such an optimizer not only uses a compute-in-memory (CIM) paradigm to address the so-called {\it memory-wall} (i.e. energy dissipated due to repeated memory \textit{read} access), but also uses a learning-in-memory (LIM) paradigm to address the energy bottlenecks due to repeated memory \textit{writes} at the precision required for optimization (the {\it update-wall}), and to address the energy bottleneck due to the repeated transfer of information between short-term and long-term memories (the {\it consolidation-wall}). In this paper, we derive theoretical estimates for the {\it energy-to-solution} metric that can be achieved by this ideal neuromorphic optimizer which is realized by modulating the energy-barrier of the physical memories such that the dynamics of memory updates and memory consolidation matches the optimization or the annealing dynamics. The analysis presented in this paper captures the out-of-equilibrium thermodynamics of learning and the resulting energy-efficiency estimates are model-agnostic which only depend on the number of model-update operations (OPS), the model-size in terms of number of parameters, the speed of convergence, and the precision of the solution. To show the practical applicability of our results, we apply our analysis for estimating the lower-bound on the energy-to-solution metrics for large-scale AI workloads. 
\end{abstract}

\maketitle


\section*{\label{sec:Introduction}Introduction}

\begin{figure*}[ht]
\centering
 \includegraphics[width=0.85\linewidth]{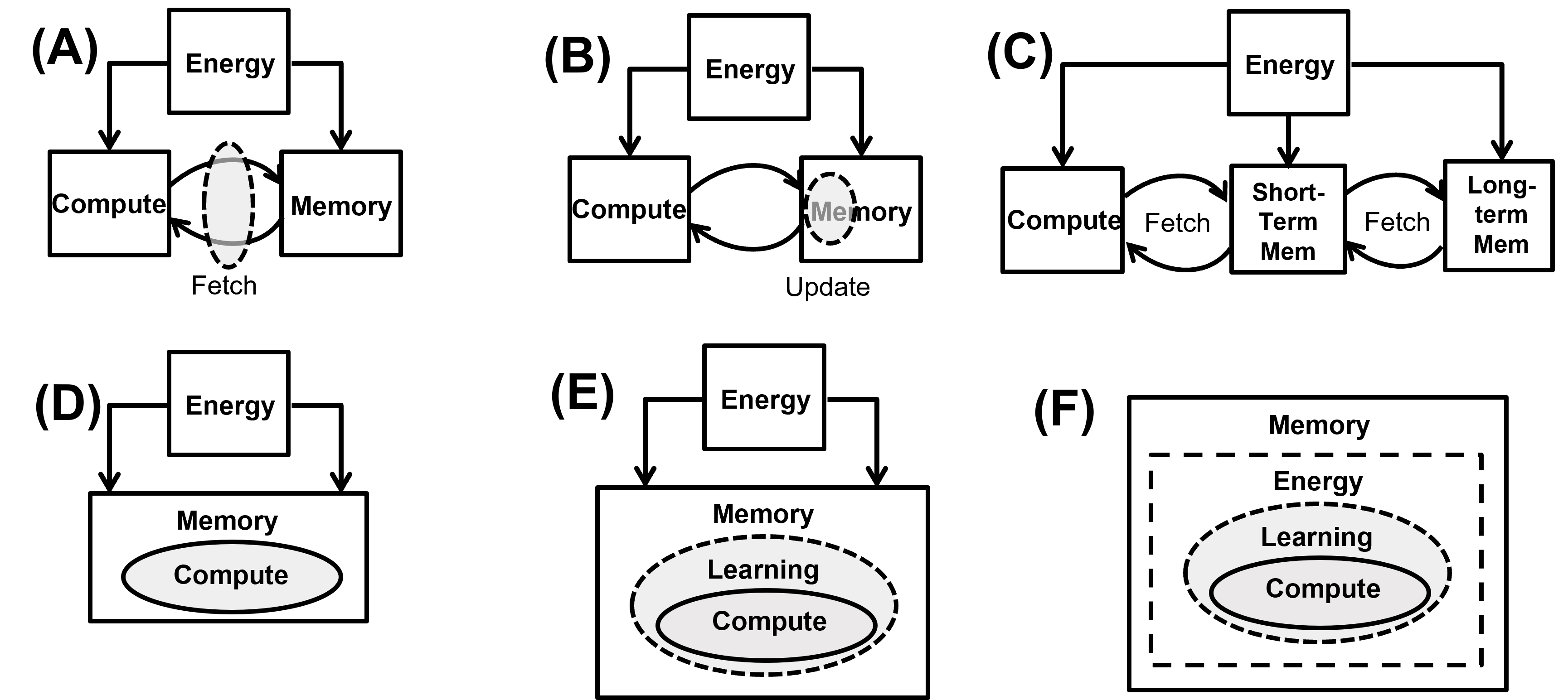}
\caption{\textbf{Abstraction of an ideal neuromorphic machine that addresses learning/training bottlenecks due to memory accesses and updates:} (A) \textit{Memory-wall} which arises due fetching data from memory units that are physically separated from the compute units; (B) \textit{Update-wall} which arises due to the frequency and precision of memory writes; and (C) \textit{Consolidation-wall} which arises due to limited memory capacity and fetching data across memory hierarchies. (D) \textit{Compute-in-memory} paradigm to address \textit{memory-wall} by co-locating memory and compute functions; (E)-(F) \textit{Learning-in-memory} paradigms  to address \textit{update-wall} and \textit{consolidation-wall} where learning is also integrated within the memory either driven by an external energy source (E) or driven by non-equilibrium process of memory erasure (F). 
    }
	\label{fig_1}
\end{figure*}

Similar to many other large-scale computing tasks, the energy footprint for training AI systems is determined primarily by the bottlenecks due to memory access and updates~\cite{chen2018CIMemoryEnergy,Horowitz2014ComputingEnergyProb}. 
Training AI systems involves searching over a large set of parameters and hence requires repeated memorization, caching, and pruning. In a conventional Von-Neumann computer architecture, where the compute and memory units are physically separated from each other, frequent parameter accesses and updates across the physical memory hierarchy contribute to significant energy dissipation. The resulting memory bottlenecks can be categorized into three \textit{performance walls}~\cite{Shantanu2023iscas} namely: the {\it memory-wall}, the {\it update-wall}, and the {\it consolidation-wall}, all of which are illustrated in Fig.~\ref{fig_1}(A),~\ref{fig_1}(B), and~\ref{fig_1}(C).

The {\it memory-wall}~\cite{Horowitz2014ComputingEnergyProb} arises because of energy dissipation due to the frequent data transfers between the computation and storage units across a memory bus (Fig.~\ref{fig_1}(A)).
The {\it update-wall} arises due to large number of memory writes (which has a larger energy footprint than memory reads) and due to the precision at which the parameter or the memory needs to be updated~\cite{Colangelo2018lowPrecisionDL, Godse2018MemoryInAI} (Fig.~\ref{fig_1}(B)). 
The {\it consolidation-wall} arises due to the limited capacity of physical memory (registers and cache) that can be integrated with or in proximity to the compute units~\cite{wang2020benchmarkAIsystem, Liu2018PIM} (Fig.~\ref{fig_1}(C)). As a result, during training, only some of the parameters of large AI models can be stored or cached locally (also called working memory), whereas the majority needs to be moved to and from or consolidated off-chip. Repeated access to this off-chip memory and the consolidation overhead in maintaining a {\it working set} of active parameters~\cite{georgiou2022greenAI} across different levels of memory hierarchy dissipates a significant amount of energy. 

To address all these memory bottlenecks, the AI community has been employing and exploring different neuromorphic (or neurally inspired) approaches to design memory and compute architectures for both training and inference. For instance, in emerging AI hardware, the {\it memory-wall} is addressed by co-locating the memory and computation functional units~\cite{Chakrabartty2007SVMclassfier,Shi2011ACMbrain,Akash2021PerspectiveCIM}, as shown in Fig.~\ref{fig_1}(D), which in part is motivated by neurobiology~\cite{Zhang2020neuroComputingChip,Wan2022CIMmemristor}.  
While this compute-in-memory (CIM) paradigm can significantly improve the energy efficiency for AI inference, unfortunately, the paradigm does not address the other two performance walls in learning (or AI training) due to the energy dissipation caused by memory writes (the {\it update-wall}), nor due to data transfers across the memory hierarchy (the {\it consolidation-wall}). In fact, the reliance on non-volatile memory may exacerbate these problems as the energy dissipated during memory writes is significantly higher than the energy dissipated to read the contents of a memory~\cite{Yuan2016memoryEnergy}.

Analogous to the CIM paradigm, neuromorphic principles have also been proposed to address the update- and the consolidation-walls. To address the {\it update wall}, rather than using backpropagation based training algorithms, neuromorphic approaches use local learning rules (like Hebbian or anti-Hebbian learning, or neuromodulation), where the synaptic weights are updated using information available locally at each memory unit~\cite{lagani2021training}. These principles also underlie recurrent architectures like Hopfield networks~\cite{ramsauer2020hopfield, Hoover2023EnergyTransformer} and neuromorphic Ising machines~\cite{Chen2025Neurosa}, where energy minimization emerge from local weight dynamics. Building on this foundation, local learning has enabled the development of newer generations of associative memories and energy transformers~\cite{oconnor19SNNEP, Gangopadhyay2021GTNN, Scellier2017EP, Laydevant2024IsingEP}.
Neuromorphic principles can also relax the precision requirements of weight updates. For instance, at a fundamental level, the precision of biological synapses as storage elements is severely limited~\cite{Kandel2001biologyMemory}. 
Despite this, some computations observed in neurobiology are surprisingly precise~\cite{Abbott2004synapticComputation}, which has been attributed to a combination of massive parallelism, redundancy, and stochastic encoding principles~\cite{Bressloff2018stochasticNeuron,vonSeelen1989parallelInNeuNet}. 
In this framework, intrinsic randomness and thermal fluctuations in synaptic devices not only aid in achieving \textit{higher} precision during learning but also improve energy efficiency through noise exploitation~\cite{Neftci2016NoiseDrivenSynapse,Sejnowski2023NoisySynapse,Chen2025Neurosa}. 

The neuromorphic solution to address {\it the consolidation-wall} has been to mimic the temporal dynamics observed in biological synapses into synthetic memory devices. There is growing evidence that biological synapses are inherently complex high-dimensional dynamical systems themselves~\cite{Fusi2005FusiSynapse,Kirkpatrick2017CatastrophicForgetting} as opposed to the simple, static storage units that are typically assumed in neural networks. This is supported by experimental evidence of synaptic {\it metaplasticity}~\cite{Kennedy2013BioSynapsePlasticity0, Koch2000SingleNeuronComp}, where the synaptic plasticity (e.g. the ``ease'' of updates) has been observed to vary depending on age and in a task-specific manner. Metaplasticity also plays a key role in neurobiological memory consolidation~\cite{McClelland1995yi, McClelland2013-fe}, where short-term information stored in ``volatile'', easy-to-update memory in the hippocampus is subsequently consolidated into long-term memory in the neocortex. 
\begin{figure*}[t!]
\centering
\includegraphics[width=\linewidth]{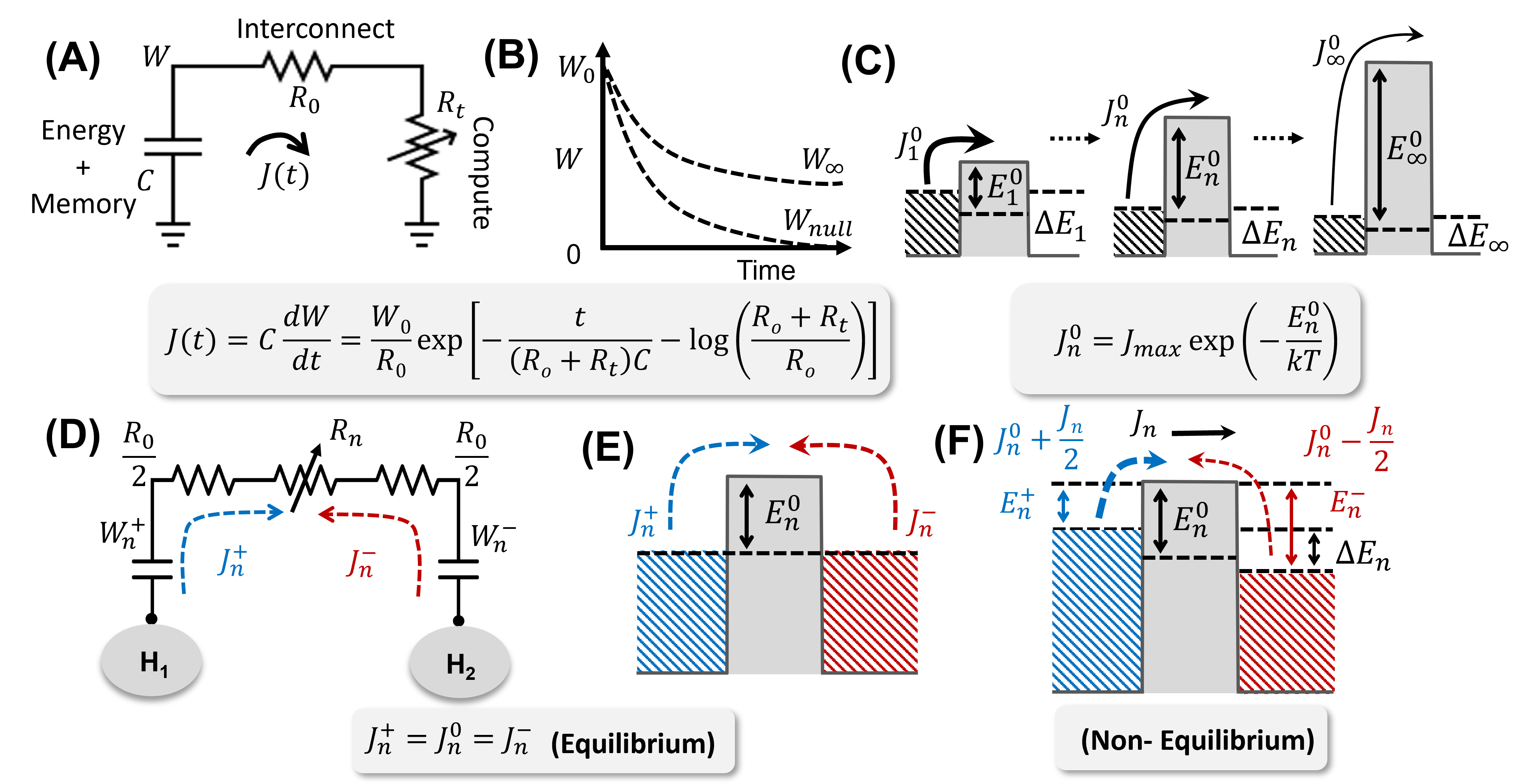}
\caption{\textbf{Abstraction of memory model for LIM: }(A) Equivalent circuit where a capacitor represents an energy-storage and a memory element, where as the compute function controls the rate of leakage through the variable resistor $R_t$; (B) Continuous-time memory decay characteristics as the value of $R_t$ is modulated; (C) A discrete-time model of the memory decay based on the modulation of the energy-barrier; (D) An equivalent circuit model to allow for bidirectional memory updates of a parameter stored as the differential voltage $ W_n = W^+_n-W^-_n$, where the differetial nodes are connected to thermal reservoir to allow current transmission; (E) The energy-band diagram representation of the differential memory model in equilibirum state where no meaningful updates of the memory is made; (F) The net memory update rate $J_n$ dissipates power when external perturbation $\Delta E_n$ is injected to the circuit to break the equilibrium. }
\label{fig_2}
\end{figure*}

Therefore, an abstraction of an ideal neuromorphic optimization engine that could address all the performance walls, would be an associative memory architecture where the compute, update, and consolidation functions are all integrated within a network of memory devices. This is illustrated in Fig.~\ref{fig_1}(E) and (F) where like compute-in-memory architectures, learning is also integrated within the network of memory devices. 
This ideal abstraction, referred to as \textit{learning-in-memory} (LIM)~\cite{Mehta2022FNsynapse,Rahman2023DynamicFNMem,Shantanu2023iscas} is the basis of our theoretical investigation in this paper where we analyze the energetics of LIM architectures, as shown in Fig.~\ref{fig_1}(E) and (F). In the LIM architecture shown in Fig.~\ref{fig_1}(E), an external energy source is used to guide the process of learning, where as the LIM architecture in Fig.~\ref{fig_1}(F), the energy dissipation due to memory erasure drives learning. Both these LIM concepts can be understood using a simple R-C equivalent circuit shown in Fig.~\ref{fig_2}(A). The charge stored on the capacitor $C$ acts as a source of energy as well as the parameter stored in the memory. During the process of learning, the capacitor discharges through a resistive network comprising of an interconnect resistance $R_0$ and a variable resistance $R$. For a fixed value of $R$, the voltage on the capacitor $W$ decays exponentially from an initial value of $W_0$ to zero, shown by $W_{null}$ in Fig.~\ref{fig_2}(B). However, if $R$ is modulated such that $R \rightarrow \infty$, then $W \rightarrow W_{\infty} \ge 0$, and the capacitor retains its charge (up to a certain precision). The energy to modulate $R$ can be provided externally which is aligned with the LIM concept in Fig.~\ref{fig_1}(E). Otherwise, $R \rightarrow \infty$ due to the inherent non-linear dynamics of $R$, which corresponds to the LIM concept in Fig.~\ref{fig_1}(F). The theoretical analysis in this paper estimates the energy-dissipated in updating the state of any physical memory (parameter update) while capturing the factors that determine memory retention (parameter storage), and combining the abstraction with the dynamics of learning and consolidation. In this regard, we believe that the results presented here are unique compared to other theoretical frameworks which focus primarily on computing bottlenecks and not on memory bottlenecks.

\section*{Results}
  
\subsection{LIM and Energy-Barrier Modulation}
We first create an abstraction of an {analog} memory model using an RC circuit in Fig.~\ref{fig_2}(A) before generalizing it to other synaptic devices.  
{We emphasize that the RC circuit is not intended to capture the underlying physics, like metastability, of a typical memory device with well-defined energy barriers, but to illustrate two essential ingredients of any physical memory: (i) the presence of a stored state variable, and (ii) a dissipative path through which that state may be read out or modified. The variable resistance element provides an intuitive way to visualize how a dynamical energy barrier could modulate the rate of state evolution. Note that the discrete states in an analog memory arise because: (a) the stored values can only be differentiated from each other up to a certain precision, due to measurement artifacts; or (b) the inherent discreteness of the memory states, which could arise due to finite number of electrons in floating-gate memory~\cite{Mehta2022FNsynapse, Rahman2023DynamicFNMem}, or due to micro-filaments acting as memory states in a memristor~\cite{Wan2022CIMmemristor}}
In the RC-model, the capacitor $C$ acts as the memory element where the parameter stored in the memory at any instant of time $t$ is represented by the charge stored or the capacitor voltage $W(t)$. The parameter changes with time as stored charge discharges through a series of resistors $R_0$ and $R_t$ that model the interconnect and leakage resistance. Note that inductive effects are ignored in this derivation. If $R_0$ and $R_t$ are assumed to be constant, the voltage $W(t)$ decays from its initial value $W_0$ as 
\begin{equation}
 W(t) = W_0 \exp\left[-\frac{t}{(R_0+R_t)C}\right]\label{eq_main_WtRC_continuous}
 \end{equation}
resulting in the a discharge current current $J(t)$ given by 
\begin{equation}
\begin{aligned}
J(t) &= -C\frac{dW}{dt} =\frac{W_0}{(R_0+R_t)} \exp\left[-\frac{t}{(R_0+R_t)C}\right] \\
 &=\frac{W_0}{R_0} \exp\left[-\frac{t}{(R_0+R_t)C} - \log\left(\frac{R_0+R_t}{R_0}\right)\right]
\end{aligned} \label{eq_main_Jt}
\end{equation}
as shown in Fig.~\ref{fig_2}(A)-(B). Note that if $R_t \rightarrow \infty$, then $J(t) \rightarrow 0$, as depicted in Fig.~\ref{fig_2}(B). However, if $R_t$ is arbitrarily modulated with respect to time, then obtaining a close-form solution of $J(t)$ is not always feasible. To make the analysis tractable, we discretize the modulation steps in intervals of $\Delta t$, and abstract the discrete time current $J_n^0 = J(n\Delta t)$ according to 

 \begin{equation}
 J_n^0 = J_{max} \exp\left(-\frac{E_n^0}{kT}\right)
 \label{eq_main_Jn0}
 \end{equation}
{where $J_{max}$ is the maximum switching rate that is specific to different hardware substrates.} {A time-varying $E_n^0$ that modulates the update rate $J_n^0$ is functionally isomorphic to the variable resistance $R_n$ affecting the charge decay current in Fig.~\ref{fig_2}(A)}. This is depicted in Fig.~\ref{fig_2}(C) where the energy-barrier height $E_n^0$ keeps increasing with time (as $R_n \rightarrow \infty$) and accordingly $J_n^0$ decreases with time. 
{$J_n^0/J_{max}$ can be viewed as a rate of transition (or escape) probability, a notion commonly used in Markovian dynamics}~\cite{peliti2021stochasticbook2, seifert2008stochasticbook1} {and in stochastic thermodynamics}~\cite{PhysRevX.10.031005, Wolpert_2020, brillouin1953negentropy}. Equation~\ref{eq_main_Jn0} {also uses coarse-graining procedure (like adiabatic elimination) from the stochastic thermodynamics literature}~\cite{peliti2021stochasticbook2} {to ignore the fast-dynamics at the top of the energy-barrier and due to the reservoir (the sink). }
Eq.~\ref{eq_main_Jn0} can also be used to estimate the height of the energy barrier required to achieve a specific reliability in storage in steady-state, as specified by a precision variable $\delta = J_{\infty}^0/J_{max}$, and is given by $E_{\infty}^0 = kT \log(1/\delta)$~\cite{Bennett1982}. In traditional AI systems using a conventional memory, this energy barrier is generally chosen to be high enough to prevent parameter leakage due to thermal fluctuations so that the parameter is retained in the memory throughout the entirety of training. For instance, in RRAM devices, where the non-volatile state of the conductive filament between two electrodes determines the stored analog value~\cite{Akinaga2010ReRam}, the energy barrier height can be as high as \SI{1}{\pico\joule}~\cite{Xue2020ReRAMCIM}. In charge-based devices like floating-gates or FeRAM, where the state of polarization determines the stored analog value~\cite{Chakrabartty2007SVMclassfier, Merrikh-Bayat2017MixedSigNeuroMem, Dünkel2017fefetCIMMVM}, the energy barrier is typically around \SI{10}{\femto\joule}~\cite{Dünkel2017fefetCIMMVM}. Therefore, an learning algorithm that adapts the stored weights in quantized steps $(\ldots, W_{n-1}, W_{n}, W_{n+1}, \dots)$ so as to minimize some system-level loss-function consumes energy to overcome the energy-barrier $E_{\infty}^0$ for each of the parameter/memory updates. In LIM the energy barrier height $E^0_n$ that retains the learned parameter can be adapted, as shown in Fig.~\ref{fig_2}(C) and it was shown~\cite{Mehta2022FNsynapse} that by matching the memory retention rates to the process of weight decay used in ML training energy efficiency could be significantly improved.

The memory model in Eq.~\ref{eq_main_Jn0} can be generalized to support differential updates (incremental and decremental) using an equivalent circuit model shown in Fig.~\ref{fig_2}(D). Here the parameters are stored as the difference between the voltages $W^+-W^-$ on the capacitors and the equivalent discharge currents are $J^+_n$ and $J^-_n$. Note that the net current flowing through the resistor $R_n$ is $J^+_n - J^-_n$. The capacitors are coupled to voltage sources $H_{1}$ and $H_{2}$, acting as equivalent chemical reservoirs and can shift the voltage levels $W^+$ and $W^-$ with respect to each other. {We assume that $H_{1}$ and $H_{2}$ are ideal voltage sources so the potential remains strictly constant during energy exchange. Energy transfer between the reservoirs and the capacitors is fully reversible, producing no entropy in the bath. Note that while the currents $J^+_n$ and $J^-_n$ can flow between the reservoirs, energy is only dissipated through the resistors and is a function of the difference between the currents.} {The bi-directional circuit analogy can similarly be mapped to the bi-potential band-diagram as shown Fig.~\ref{fig_2}(E), where the energetic difference across the energy barrier $E^0_n$ encodes the information stored. When no content is stored in the memory, the system is in equilibrium as} 
 \begin{align}
    J^+_n &= J^-_n=J^0_n.\label{eq_RC_current_balanced}
\end{align}
{where the update rate in both directions are equal to each other, and thus no energy is dissipated. When memory is updated as shown in Fig.~\ref{fig_2}(E), an external field  biases the energy landscape so that the top of the energy-levels on both sides no longer lie at equal energy. This energy asymmetry establishes a preferred state, which constitutes the stored information. Once the memory is updated, the effective finite barrier height $E^+_n$ and $E^-_n$ across the barrier $E^0_n$ determines the rate of thermally activated transitions and therefore the retention and update dynamics}
\begin{equation}
\begin{aligned}
    J_n^+ &= J_n^0 + \frac{J_n}{2}  \\
    J_n^- &= J_n^0 - \frac{J_n}{2},
\end{aligned} \label{eq_main_current_differential}
\end{equation}
where $J_n$ denotes the net update rate across the barrier. Also note that as $E^0_n \rightarrow \infty$, $J^+_n,J^-_n \rightarrow 0$; hence similar to Fig.~\ref{fig_2}(A) and (B), the energetic difference across the barrier $\Delta E$ could be non-zero and encodes the stored parameter, {which in turn causes the energy dissipation in the process of memory writes}. Then, assuming that the chemical reservoir is always at equilibrium~\cite{peliti2021stochasticbook2,seifert2008stochasticbook1} at time-scales of the memory updates, $J_n$ (which represents probability flow) can be written as
\begin{equation}
\begin{aligned}
    J_n &= J_{max}\cdot\left[\exp \left(-\frac{E^+_n}{kT} \right)-\exp \left(-\frac{E^-_n}{kT}. \right)\right]   
\end{aligned} \label{eq_main_current_dE}
\end{equation}
which leads to
\begin{figure*}[ht!]
\centering
\includegraphics[width=0.95\linewidth]{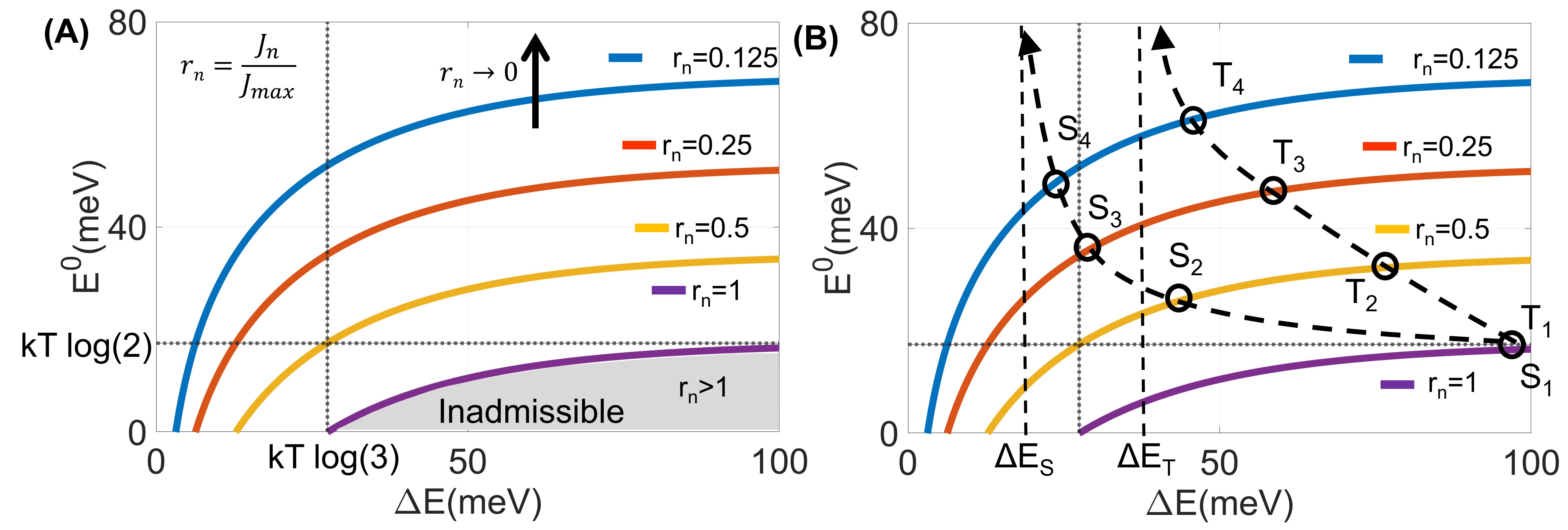}
\caption{\textbf{The energy-barrier vs energy-difference(stored parameter) or $E^0_n$-vs-$\Delta E_n$ plot:} (A) Retention (energy barrier height) versus stored parameter for different normalized update rates $r_n = \frac{J_n}{J_{\text{max}}}$ that can achieved for a given $E^0$ and $\Delta E$. The shaded area indicates the region of inadmissible update rates. (B) Different LIM algorithms where memory elements follow different trajectories to the final steady-state solutions, denoted as $\Delta E_S$ and $\Delta E_T$ respectively. As the energy gradient is minimized while the retention increases, both factors contribute to the asymptotically decreasing update rate $J_n$. 
 }
\label{fig_3}
\end{figure*}

\begin{equation}
\begin{aligned}
    J_n= 2 J_{\text{max}} \exp \left( \frac{-E_n^o}{kT} \right) 
    \left[\frac{\exp \left( \frac{-\Delta E_n}{kT} \right)-1}
    {\exp \left( \frac{-\Delta E_n}{kT} \right)+1}\right]
\end{aligned}\label{eq_main_Jn2}
\end{equation}
and
\begin{align}
    E_n^0 &= kT \log \left[\frac{2J_{\text{max}}}{J_n}\cdot\frac{\exp\left(-\frac{\Delta E_n}{kT}\right)-1}{\exp\left(-\frac{\Delta E_n}{kT}\right)+1}\right].\label{eq_main_master}
\end{align}
This relationship between $J_n$, $\Delta E_n$ and $E^0_n$ provides insight into how the barrier height ($E_n^0$) should change, or in turn, how the resistor $R_n$ should be modulated for a given memory update rate $J_n$ and external energy $\Delta E_n$.

In Fig.~\ref{fig_3}, we plot Eq.~\ref{eq_main_master} for different values of the normalized update rate $r_n=\frac{J_n}{J_{\text{max}}}$. Along the $x$-axis the energy barrier is absent (i.e. for $E_n^0 = 0$) and the memory updates are driven directly by the presence of $\Delta E_n$. Along the $y$-axis the stored parameter reaches the trivial solution (i.e. $\Delta E_n = 0$), hence the memory update rate is zero. {Note that in Eq.~\ref{eq_main_master}, in the limiting regime where the update rate approaches its maximal value ($r_n \!\to\! 1$) and the energetic bias between states becomes very large ($\Delta E_n \!\to\! \infty$), the characteristic thermal energy scale associated with the minimum barrier height reduces to $E_n^0 = kT\ln 2$. This scale reflects the fact that, in a thermally fluctuating environment, two memory states must be separated by a free-energy difference on the order of $kT$ in order to remain reliably distinguishable over the relevant timescale. The value $kT\ln 2$ coincides numerically with the Landauer bound for logically irreversible bit erasure~\cite{Landauer1961}. It should be interpreted only as a consistency check on the underlying thermodynamic energy scale, rather than as a statement that the present analog retention process constitutes Landauer erasure.}
Fig.~\ref{fig_3}A also shows the infeasible/inadmissible region where $r_n=\frac{J_n}{J_{\text{max}}}>1$, i.e. where $J_n$ exceeds the maximum update-rate $J_{\text{max}}$.
During the process of learning, $\{(E_n^0,\Delta E_n)\}_{n=1}^\infty$ determines the computation rate $\{J_n\}_{n=1}^\infty$ according to the Eq.~\ref{eq_main_master}. 
Asymptotically, as $n \rightarrow \infty$, the learned parameters need to be retained, therefore memory leakage must be reduced by increasing the height of the energy barrier $E_n^0$, or $J_n$ must go to zero. To achieve this, different LIM algorithms will follow different trajectories in the $E^0_n$-vs-$\Delta E_n$ plot, as shown by $S_1,S_2,\ldots$ and $T_1,T_2,\ldots$ in Fig.~\ref{fig_3}(B). Asymptotically, trajectories $S$ and $T$ converge to $\Delta E_S$ and $\Delta E_T$, which represents the learned parameters (similar to the capacitor voltage in Fig.~\ref{fig_2}(B)).

\begin{figure}[ht!]
\centering
\includegraphics[width=0.95\linewidth]{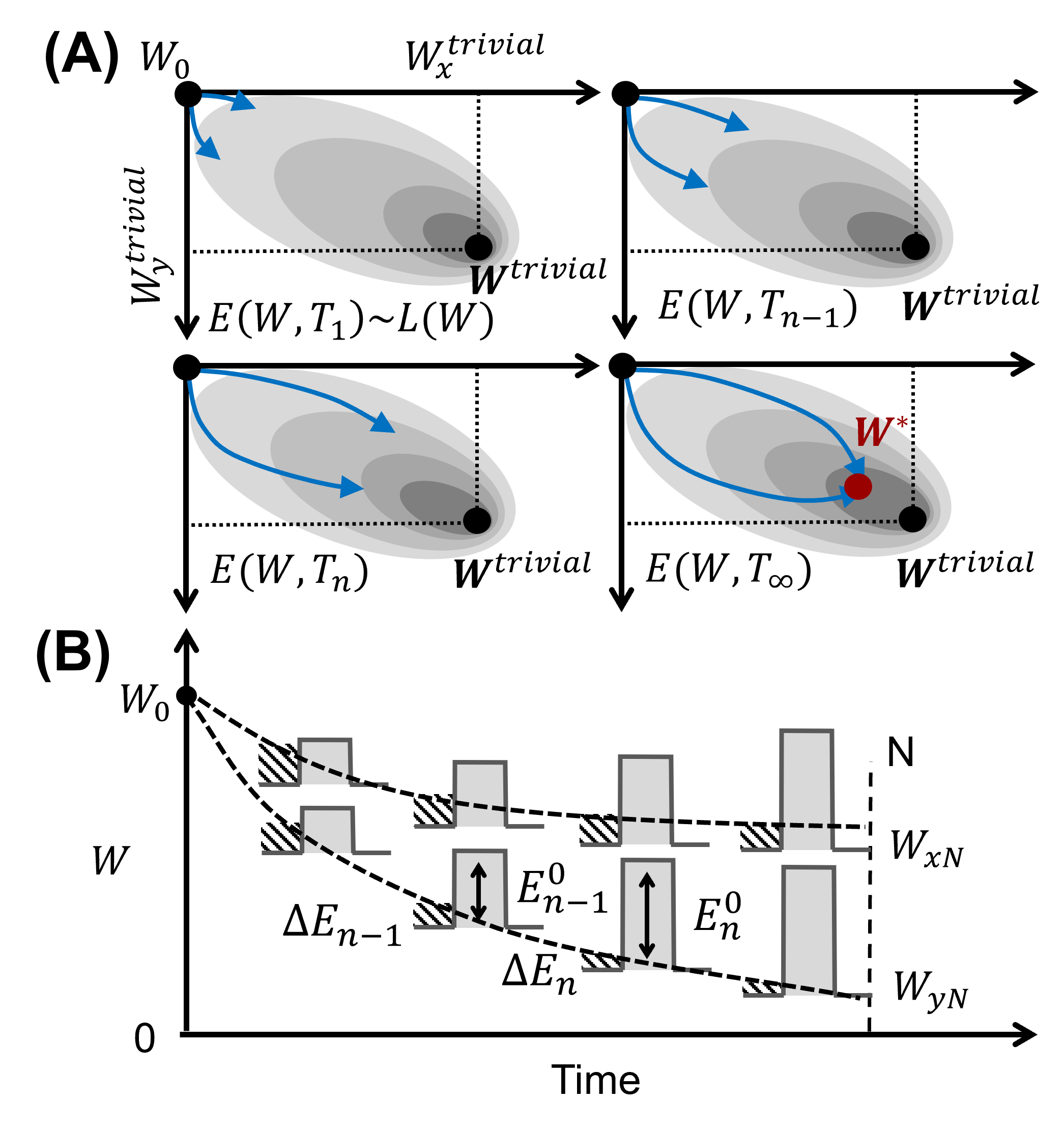}
\caption{\textbf{Illustration of LIM for a two-parameter model:} A. With a time-varying energy barrier($E^0_n$) of the memory system, the energy dynamics of the system varies w.r.t. time $T_n$, where the initial system energy $E(W, T_1)$ corresponds to the learning objective/loss function $L(\mathbf{W})$. 
For a two-parameter LIM system, $W_{x,n}$, $W_{y,n}$ evolves according to the instantaneous energy contour of the system $E(W,T_n)$ at time $T_n$. The barrier modulation dynamics are designed such that the learned parameter converges to the optimal solution $\it{\mathbf{W}^*}$ and avoid the trivial solution $\it{\mathbf{W}^{trivial}}$ for the learning objective $L(\mathbf{W})$. 
B. The parameters $W_x$ and $W_y$ overcome dynamically modulated energy barrier in LIM memory in discrete time steps $n$.}
	\label{fig_4}
\end{figure}

Fig.~\ref{fig_4}(A) demonstrates the dynamics of an LIM system with two parameters ($W_x$ and $W_y$) where learning involves minimizing an objective/loss function $L(\mathbf{W})$. Like the R-C example in Fig.~\ref{fig_2}(B), if the energy-barrier $E_n^0$ is held constant, the memory storing the parameters would decay from the initial state ${\bf W}^0$ to the trivial solution $\mathbf{W}^{trivial}$ following a fixed system energy contour $E(\mathbf{W})$ equivalent to $L(\mathbf{W})$. As shown in Fig.~\ref{fig_4}(A), at a finite time $T_n$ the energy-barrier modulates the update rate so that the effective energy contour $E(\mathbf W, T_n)$ is flat near the final solution. As a result, the memory updates would stop at a different state ${\bf W}^*$ and avoid decaying to the trivial solution $\mathbf{W}^{trivial}$. 
Note that the process of memory update (or $J_n$) is driven by the process of entropy production where ${\bf W}^0$ represents the initial low-entropy state. Hence in LIM, the process of learning (or memory updates) could follow different possible trajectories $T_{LIM}$ to the desired solution ${\bf W}^*$. The energy required to guide the process is used for modulating the respective energy barriers $E_n^0$ and the energetic difference $\Delta E_n$ that encodes the stored parameters, according to the Eq.~\ref{eq_main_master}. Also for practical reasons, the final asymptotic (or retained) state needs to be within the neighborhood of ${\bf W}^*$ (to some predetermined precision). In the next section, we show the specific LIM trajectory could be chosen according to optimal memory consolidation dynamics. 

\subsection{LIM and consolidation trajectory}\label{sec_res_grad_trajectory}

We consider a generalized learning model that comprises of $M$ parameters represented by a vector ${\bf W} \in \mathbb{R}^M$. At a time-instant $n$, the $d^{th}$ parameter $W_{d,n}$ is updated according to the standard weight-decay as
\begin{equation}
	W_{d,n} =\eta_n W_{d,n-1}-\epsilon_ng_{d,n}
	\label{eq_main_weightdecayGD}
\end{equation}
where $\eta_n=(1-\lambda\epsilon_n)$ denotes a weight-decay parameter~\cite{Loshchilov2017AdamW, Hanson1988WeightDecayRegularizer}, $\epsilon_n$ denotes the learning-rate and $\lambda$ a hyperparameter that controls the rate of the decay process. The term $g_{d,n}$ in equation~\ref{eq_main_weightdecayGD} denotes a reinforcement parameter which could simply be the gradient of a loss-function. Unrolling the recursion in equation~\ref{eq_main_weightdecayGD}, $W_{d,n}$ can be expressed as
\begin{equation}
	W_{d,n} =P_{1:n} W_{d,0}-\sum_{k=1}^nP_{k+1:n}\epsilon_kg_{d,k},
	\label{eq_main_weightdecayGD1}
\end{equation}
where $P_{l:n}$ denotes $\prod_{k=l}^n\eta_k=\prod_{k=l}^n(1-\lambda\epsilon_k)$ and $W_{d,0}$ denotes the initial weight values. 
Without any loss of generality, we will assume that the $M$ parameters $W_{d,n}$ are centered around zero (mean removed). Therefore, the average $\Delta E_n$ across all the $M$ parameters is proportional to the variance $\sigma^2(W_{d,n})$ according to
\begin{equation}
	\Delta E_n = \beta kT \sigma^2(W_{d,n}). \label{eq_main_dE2dL}
\end{equation}
where $\beta > 0$ and $kT$ are the conversion factors that has been introduced to convert algorithmic factors into physical factors. To simplify our analysis, we assume that the variance on the initial weights $\sigma^2(W_{d,0})$ can be ignored (due to deterministic initialization). We will use the assumptions described in the memory consolidation literature~\cite{Fusi2005FusiSynapse, Benna2016FusiConsolidation}, where $g_{d,n}$ are assumed to be uncorrelated with respect to each other in time and its variance is constant $\sigma_g^2$ (special case of signed update). {It should be noted that this assumption is a common practice in theoretical machine learning~\cite{Robbins1951ASA} to understand the asymptotic convergence and for analytic tractability. However, in practice, many learning algorithms violate this assumption as described in the Discussion section. }

{Following Eq.~\ref{eq_main_dE2dL}} 
\begin{equation}
        \Delta E_n = \beta kT \sigma_g^2 \sum_{k=1}^nP^2_{k+1:n}\epsilon_k^2.
	\label{eq_main_weightdecayGD3}
\end{equation}
Since $E_n$ needs to be bounded (to model finite memory capacity), we follow the rationale provided in~\cite{Fusi2005FusiSynapse, Benna2016FusiConsolidation} for an optimal memory consolidation schedule where $\epsilon_k$ decay is given by $\epsilon_k \sim (c/\sqrt k)$. Substituting the optimal learning rate dynamics into Eq.~\ref{eq_main_weightdecayGD3} and using the fact that $0<\lambda c\ll1$~\cite{He2015ResNet, Devlin2018BERT} we get
\begin{align}
	\Delta E_n &= \beta kT \sigma_g^2 c^2\sum_{k=1}^n\exp{\left(2\lambda c(\sqrt n - \sqrt k)\right)}\frac{1}{k}\\
    &\geq\beta kT \sigma_g^2 \frac{c^2}{n}\sum_{k=1}^n\exp{\left(2\lambda c(\sqrt n - \sqrt k)\right)}.\label{eq_main_weightdecayGD4}
\end{align}
Since the series in Eq.~\ref{eq_main_weightdecayGD4} are dominated by terms that are closer to $n$, let $\frac{p}{n}\ll 1$, $\quad p=n-k$, we can approximate $\sqrt n-\sqrt k$ using first order expansion as $\frac{p}{\sqrt n}$. Therefore, Eq.~\ref{eq_main_weightdecayGD4} can be rewritten as 
\begin{align}
	\Delta E_n &\geq\beta kT \sigma_g^2 \frac{c^2}{n}\sum_{p=0}^{n-1}\exp{\left(2\lambda c\frac{p}{\sqrt n}\right)}\\
    &\geq\beta kT \sigma_g^2 \frac{c}{2\lambda}\frac{1}{\sqrt n}\label{eq_main_weightdecayGD5}
\end{align}
The detailed derivation can be found in Appendix section~\ref{sec_supp_caliberation}, where we also show
that the constants appeared in Eq.~\ref{eq_main_weightdecayGD4} are related to the precision of the memory $\delta$ according to $\beta\sigma_g^2\frac{c}{2\lambda} = 1/\delta$. {More specifically, $\delta$ denotes the precision at which a value can be stored in the memory or can be measured. The uncertainty in retention or measurement is reflected by the probability that noise (thermal fluctuations) causes an erroneous state transition or readout. Following the thermodynamic analysis of Brillouin~\cite{Brillouin1956} and Bennett~\cite{Bennett1982}, the minimal free energy required to reliably distinguish two physical states with error probability $\delta$ is equivalent to the minimal energy barrier height the memory must exhibit, }

\begin{equation}
        E_\infty^0 \ge kT \log(\frac{1}{\delta}).
	\label{eq_main_energy_barrier_with_delta_precision}
\end{equation}
{Also, based on Shannon information, resolving $P$ bits of information requires that the uncertainty in distinguishing the stored states be at most one part in $2^P$. Therefore, the probability of misidentifying a state must be no larger than $\delta\sim2^{-P}$ and hence $\delta$ reflects the bit-precision $P$ of the model parameter.}

Note that Eq.~\ref{eq_main_weightdecayGD5} recovers the optimal learning rate dynamics $\epsilon_n$, which after combining with   Eq.~\ref{eq_main_master} leads to
\begin{align}
    & \Delta E_n \ge \left(\frac{kT}{\delta}\right) \frac{1}{\sqrt n} \quad \label{eq_energyEpsDelta-1}\\
    \text{and } & E_n^0 \ge kT \log \left[\frac{2J_{\text{max}}}{J_n} \cdot \frac{\exp\left(-1/\sqrt n\delta\right)-1}{\exp\left(-1/\sqrt n\delta \right)+1}\right] \label{eq_energyEpsDelta-2}
\end{align}

Eq.\ref{eq_energyEpsDelta-1}-\ref{eq_energyEpsDelta-2} is model agnostic and connects the \textit{precision} at which the parameters are updated and retained in memory during and at the end of training, and the resulting energy bounds. 
The update-rate $J_n$ models the speed of computation at a time-instant $n$, and $\delta$ models the precision of the memory update/retention, both of which are metrics associated with the {\it update-wall}. The learning-rate $\epsilon_n$ determines the dynamics of memory-consolidation~\cite{McClelland1995yi, Fiebig2014MemConsolidationLearningRate}, and hence is a metric associated with the {\it consolidation-wall}. It is to be noted that, the precision modeled by $\delta$ is different from the $P$ bits of precision for the gradient of loss function. Here, $\delta$ denotes the precision of the memory retention/update at the final state. Thus, Eq.~\ref{eq_energyEpsDelta-1}-\ref{eq_energyEpsDelta-2} is the key to derive and understand the energy-efficiency limits of LIM-based training.
\begin{figure*}[ht]
	\centering
 	\includegraphics[width=0.9\linewidth]{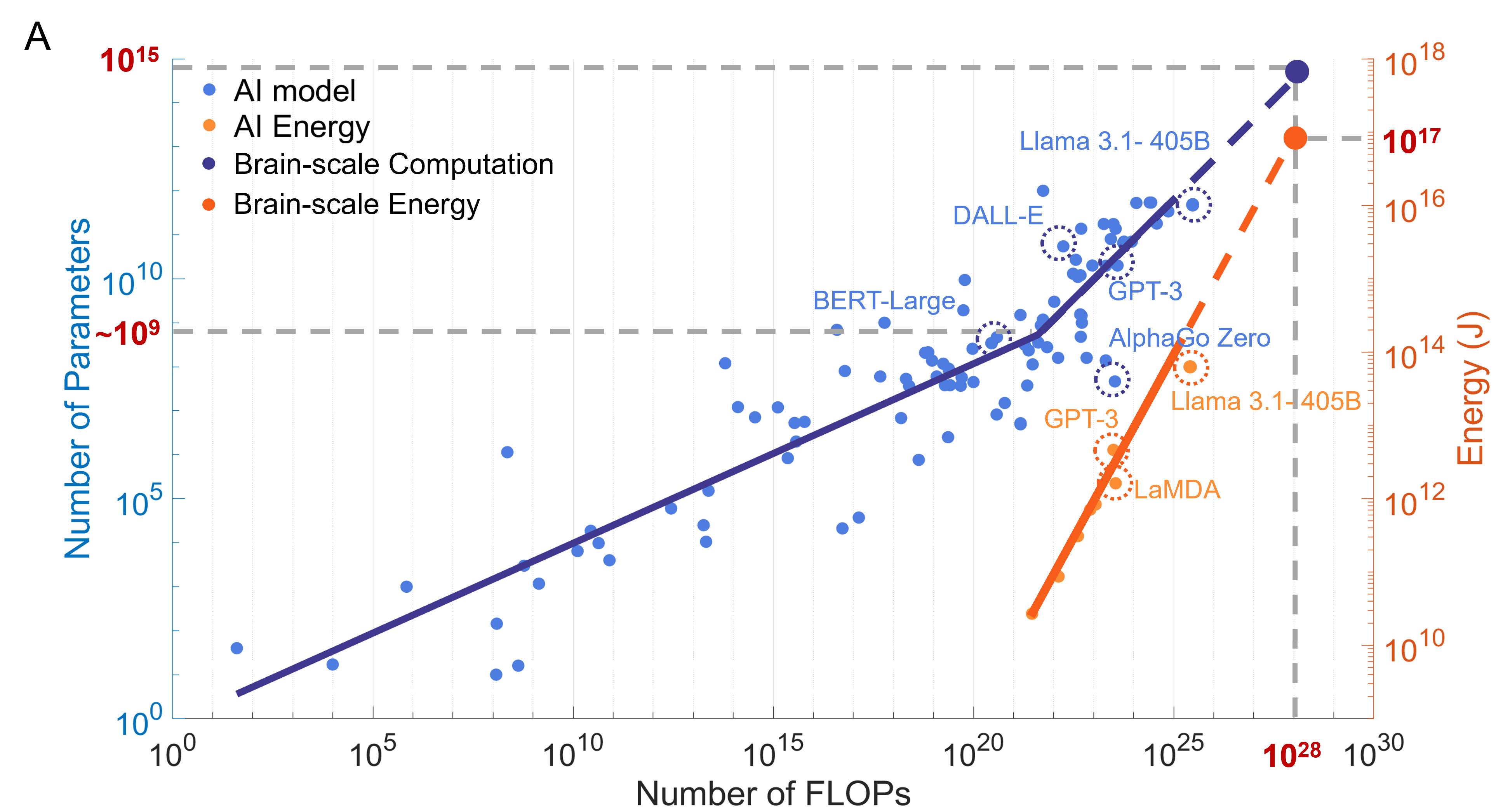}
	\caption{\textbf{Empirical data used for estimating agnostic energy-efficiency bounds:} Trends showing the growth of computational and energy needs for training AI models~\cite{Giattino2023aiData,grattafiori2024llama3herdmodels}, which has been used to predict the \num{e28} FLOPs and \SI{e17}{\joule} of energy that would be needed to train a brain-scale AI model}
	\label{fig_7}
\end{figure*} 

\subsection{Estimation of LIM energy dissipation }\label{sec_res_limvariants}
Eq.~\ref{eq_energyEpsDelta-1}-\ref{eq_energyEpsDelta-2} shows that setting any two of the three parameters learning-rate~$\epsilon_n$, update-rate~$J_n$, energy barrier-height~$\Delta E_n$, determines the third. 
This relationship can be used to estimate the total energy dissipation $E_{\text{Total}}$ for an LIM-based training. 
$E_{\text{Total}}$ is determined by the energy dissipated for building up the barrier $E_{\infty}^0$ at the end of training and the energy dissipated $\Delta E_n$ at each time instant $n$ to support the flow $J_n$.

For $M$ parameters or memory elements, the total energy dissipation $E_{\text{Total}}$ can be approximated as
\begin{equation}
 \textstyle E_{\text{Total}} \approx  M \left(kT\sum_{n=1}^{\infty} J_n \Delta E_n \Delta t  + E_{\infty}^0\right),
 \label{eq_main_Etotal_general}
\end{equation}
where $\Delta t$ denotes the duration of each discrete time step. Note that the expression for energy dissipation in equation~\ref{eq_main_Etotal_general} is aligned with the heat-flow expressions used in stochastic thermodynamics literature~\cite{peliti2021stochasticbook2}. 
We also enforce constraints to ensure: (a) termination of training within a finite number of operations; and (b) the required memory retention at the end of training. These are specified as
\begin{align}
   \textstyle M \sum_{n=1}^{\infty} J_n \Delta t =\#FLOPs \nonumber \\
   \quad\mbox{ and }\quad E_{\infty}^0 \ge kT \log (1/\delta)
   \label{eq_main_constraints_E0_flops}
\end{align}
where $\#FLOPs$ denotes the total computational workload need to complete a task. In Section~\ref{sec_res_trend}, we use empirical trends based on reported metrics to estimate the $\#FLOPs$ required. Note that unlike speed-limit theorems in stochastic thermodynamics~\cite{doi:10.1073/pnas.2321112121}, in Eq.~\ref{eq_main_constraints_E0_flops} we are enforcing a task-completion within a finite time-horizon by choosing a specific schedule on the update rate $J_n$.  
We also constrain the update-rate schedule $r_n = \frac{J_n}{J_{\text{max}}}$ to be non-increasing, between zero and one, monotonic and asymptotically bounded (i.e. $r_n \overset{n\to\infty}{\longrightarrow} 0$ and $\sum_{n=1}^\infty r_n\leq B, \quad B>0$). Note that these conditions complement the constraints imposed on update rate $r_n$ in Appendix section~\ref{sec_supp_caliberation}, where we calibrate the hyper-parameters under the the adiabatic (slowest) bound on the update rate, while in practice, constraints in Eq.~\ref{eq_main_constraints_E0_flops} enforce the total computation ($\#FLOPs$) remains bounded. Based on the constraints, a natural choice is a polynomial decay
$r_n = n^{-(1+\gamma)}$ with $\gamma \ge 0$.
This choice of $r_n$ ensures that $J_n \le J_{\text{max}}$ and $\Delta E_n \ge 0$ for all~$n$. Then  Eq.~\ref{eq_main_Etotal_general} and  Eq.~\ref{eq_main_constraints_E0_flops} can be used to eliminate intermediate variables $J_{\text{max}}$ and $\Delta t$ to produce
\begin{equation}
 E_{\text{Total}} \approx  \#FLOPs \left[\frac{\sum_{n=1}^{\infty} \Delta E_n r_n}{\sum_{n=1}^\infty r_n}\right]  + M E_{\infty}^0.
 \label{eq_main_Etotal_general_rn}
\end{equation}
Note that  Eq.~\ref{eq_main_Etotal_general_rn} incorporates two forms of energy dissipation: 
(a) the first term represents the dynamic energy dissipated during every operation; and 
(b) the second term represents the energy that is stored in the energy barriers (for memory retention).
Even though the barrier energy can potentially be recycled, for our analysis we will assume that this energy cannot be recovered.  

For LIM, the memory updates are driven by the energy stored as parameter in memory by Eq.~\ref{eq_energyEpsDelta-1} and shown in Fig.~\ref{fig_3}. We show in the Appendix subsection\ref{sec_supp_energy_estimates} that for $M$ parameters or memory elements, the total energy dissipation $E_{Total}$ for LIM under the optimal learning rate schedule $\epsilon_n\sim\frac{1}{\sqrt n}$ can be estimated as
\begin{equation}
 E_{Total} \approx  \#FLOPs \left(\frac{kT}{\delta}\right) \left[\frac{\sum_{n=1}^{\infty}  r_n/\sqrt n}{\sum_{n=1}^{\infty} r_n}\right]  + M E_{\infty}^0.
 \label{eq_main_EtotalA}
\end{equation}

While several choices of $r_n$ are possible, for the sake of exposition, we only present the results for the polynomial update-rate schedule $r_n = 1/n^{1+\eta}$ with $\gamma >0$ and the Appendix section~\ref{sec_supp_exponential_analytical} present analysis for other update-rate schedules. For the polynomial update-rate, an analytical expression for LIM energy dissipation estimates can be derived from Eq.~\ref{eq_main_EtotalA} as
\begin{equation}
    E_{Total} \approx \#FLOPs \left(\frac{kT}{\delta}\right) \left[\frac{\zeta(3/2+\gamma)}{\zeta(1+\gamma)}\right] + M kT \log \left[ \frac{1}{\delta} \right]
    \label{eq_main_EtotalA_zeta}
\end{equation}
where 
$\zeta(.)$ is the Riemann-zeta function.


\subsection{Numerical estimation of LIM energy-dissipation for AI workloads}\label{sec_res_trend}

\begin{figure*}[t]
\centering
\includegraphics[width=1\linewidth]{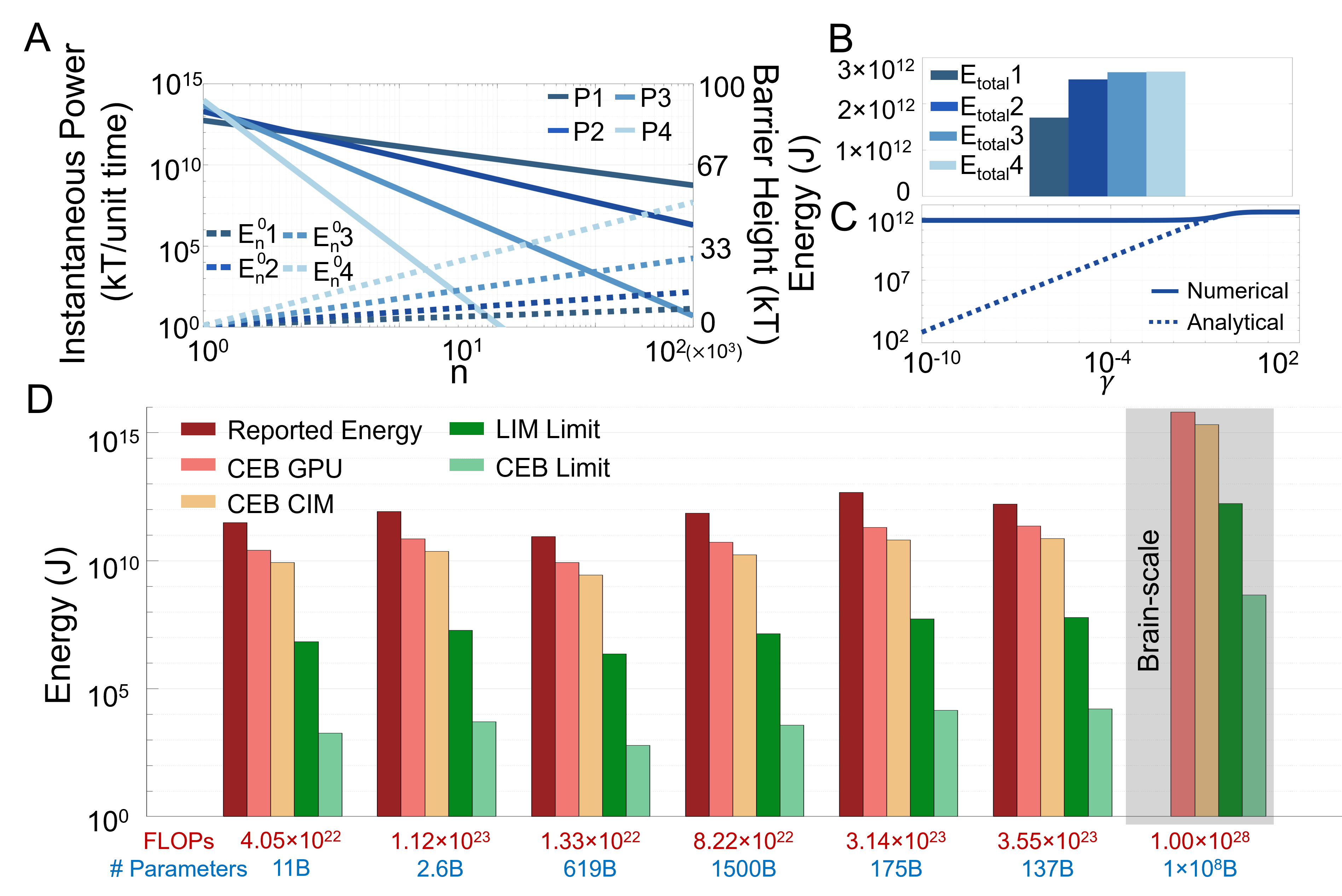}
\caption{\textbf{Energy dissipation estimates when training under LIM paradigm}(A) Estimated instantaneous power dissipation $P$ and energy barrier height $E_n^0$ for training brain-scale AI using LIM lower-bounds with a polynomial update rate $J_n\sim\frac{1}{n^{1+\gamma}}$ with $\{P1,E_n^01\sim\gamma=0.5;P2,E_n^02\sim\gamma=2;P3,E_n^03\sim\gamma=5;P4,E_n^04\sim\gamma=10\}$; (B) Total energy consumption of LIM under update rate schedules in (A): $E_{total}1 = $\SI{1.70e12}\joule; $E_{total}2 = $\SI{2.53e9}\joule; $E_{total}3 = $\SI{2.68e12}\joule; $E_{total}4 = $\SI{2.70e12}\joule; (C) Total LIM energy dissipation against update rate hyperparameter $\gamma$ in discrete finite simulation and infinite Riemann-zeta forms ; (D) Estimated energy dissipated compared against the reported energy dissipation for real-world AI training workloads~\cite{Gonzalez2021TransformerEnergy}. From left to right, the corresponding AI models are T5, Meena, Gshard-600B, Switch Transformer, GPT-3, and LaMDA. The Brain-scale AI model estimates is for a precision $\delta = 2^{-16}$. LIM$_A$ estimate is based on Eq.~\ref{eq_main_EtotalA}, and estimate of CEB limit is based on Eq.~\ref{eq_main_Etotal_general}.}
\label{fig_8}
\end{figure*}

To ensure that the energy estimates are agnostic to the idiosyncrasies of a learning algorithm and learning heuristics, we abstract the problem of AI training in terms of parameters that are common to all learning and optimization methods. These parameters include model size (or number of parameters stored in memory), the number of operations required to train the model, the learning-rate schedule, and the rate and precision of memory/parameter updates. The relationship between model size (number of parameters) and computation complexity (number of FLOPS) can be extrapolated from numbers reported in the literature. This is shown in Fig.~\ref{fig_7}, which plots the relationship between the reported number of floating-point operations (FLOPS) required to train AI systems of different sizes and the number of trainable parameters. For smaller models (number of parameters less than $1$B), the number of training floating-point operations (\#FLOPS) grows quadratically w.r.t. the number of model parameters. This trend becomes linear for larger AI models like large language models (LLMs), as shown in Fig.~\ref{fig_7}. One speculation for this phase transition at model size of $1$B is that the parameters could be directly stored in the main memory and as a result, the energy cost of optimal pair-wise comparison is manageable. Beyond $1$B parameters, external storage needs to be accessed, in which case the prohibitive energy cost dictates practical online training algorithms whose complexity grows linearly.
Assuming that this linear trend will continue for larger models, it can be projected that the number of FLOPS required to train a brain-scale AI system comprising 1 quadrillion parameters ($10^{15}$, roughly the number of synapses in a human brain~\cite{HerculanoHouzel2009brainSize,HerculanoHouzel2012brainScale}) would be around \num{e28}, as shown in Fig.~\ref{fig_7}.

To estimate the total energy that might be required to train such a system, we use another trend that relates the energy consumption to the number of FLOPS used during training. 
The trend shown in Fig.~\ref{fig_7} appears to be linear and is benchmarked against reported energy dissipation metrics for several AI systems~\cite{Gonzalez2021TransformerEnergy,DeFreitas2022LaMDA,Brown2020GPT3}. 
By extrapolating this FLOPs-to-energy relation to a brain-scale AI model one can estimate the energy required to train the model to be \SI{e17}{\joule} or equivalently \SI{2.78e7}{\mega Wh}. 

Now we shift focus to explore the fundamental energy bound for training such workloads under our LIM theoretical framework. Fig.~\ref{fig_8}(A) plots the energy barrier-height $E_n^01-E_n^04$ and the estimated instantaneous power $\Delta E_n J_n$ with different update-rate schedules $P1-P4$. Each schedule corresponds to a different value of $\gamma$ in $r_n = 1/n^{1+\gamma}$. Note that for this experiment we have used the numerical expression given by Eq.~\ref{eq_energyEpsDelta-2} for parameters $\#FLOPS$ and $M$ that correspond to a brain-scale AI workload (extrapolated from Fig~\ref{fig_7}). For the polynomially decreasing update rate, the barrier height increases logarithmically. As a result, the instantaneous power dissipation for LIM systems decreases monotonically. Also note that at the end of the training $E^0_{\infty}$ is higher and satisfies the constraint~\ref{eq_main_constraints_E0_flops}.
 
Fig.~\ref{fig_8}(B) plots the total energy dissipated for LIM system to train a brain-scale ML model, as given by Eq.~\ref{eq_main_EtotalA} for $4$ different update rate dynamics in Fig.~\ref{fig_8}(A). Note that the total energy is equivalent to estimating the area under the curve (AUC) of the instantaneous power plots in Fig.~\ref{fig_8}(A). To understand how the hyperparameter $\gamma$ in the update rate schedule affects the total energy dissipation, in Fig.~\ref{fig_8}(C), we plot $E_{total}$ against $\gamma$. As evident in the figure, in discrete finite time simulation according to Eq.~\ref{eq_main_EtotalA}, the energy dissipation of LIM system is encapsulated in an envelope of [\SI{5.81e11}, \SI{2.70e12}]. Whereas the infinite Riemann-zeta sum from Eq.~\ref{eq_main_EtotalA_zeta} shows that the energy consumption goes down indefinitely with the update rate approaching the adiabatic limit, validating the experimental finds in~\cite{Brut2012LandauerFiniteTime, Zhen2021Universal_bound} that the energy consumption for computation approach the adiabatic(Landauer's Limit) as the rate decreases. However, since the memory update is an atomic operation, the update needs to stop within finite time constraint, in numerical simulation we adopt the lower bound of the envelope in the discrete finite time simulation as the practical energy lower bound for LIM with learning rate $\epsilon_n\sim\frac{1}{\sqrt n}$ and polynomially decaying update rate $r_n\sim\frac{1}{n^{1+\gamma}}$.


Next, we estimate the LIM energy-dissipation for realistic AI workloads based on the reported metrics (training $\#FLOPS$ and number of training parameters) as shown in Fig.~\ref{fig_7}.
For comparison, we also estimated the energy dissipation for the same workloads using reported metrics for graphical processing units (GPUs), and resistive RAM(RRAM) CIM architecture~\cite{Wan2022CIMmemristor}. The energy estimation procedure is described in the Appendix subsection \ref{sec_material_convmem}. For further reference, we also include the measured energy dissipation reported in the literature~\cite{Gonzalez2021TransformerEnergy}. As shown in Fig.~\ref{fig_8}(C), the reported energy is within $1$ order of magnitude of the optimistic energy estimates using GPU statistics as described in Eq.~\ref{eq_supp_totalEnergyGeneral}, validating the effectiveness of our approach for estimating the total energy consumption from the model parameters, i.e. parameter size and total FLOPs. On the other hand, the energy lower bound for training the same AI workload using the LIM system are $7$ orders of magnitude lower than the reported energy and 4 to 5 orders of magnitude lower than the optimistic estimates of GPU and CIM platforms. Also, LIM estimates are higher than the consant energy barrier(CEB) limit, where the barrier height remains constant and given by the Landauer's limit under the target precision. This is exptected because Landauer's limit dictates that the memory updates proceed at the adiabatic rate under the precision requirement. Therefore, LIM systems with a finite update rate and termination schedule would dissipate more power than the CEB energy limit~\cite{Brut2012LandauerFiniteTime, Zhen2021Universal_bound}. 

\section*{Discussion}\label{sec_discussion}
In this paper we presented the energy-dissipation estimates for LIM that are model-agnostic and depends only on the training $\#FLOPS$, the number of model parameters $M$, the precision of memory retention $\delta$ and a hyperparameter $\gamma$ that models the decay in the update-rate schedule. In particular Eq.~\ref{eq_energyEpsDelta-2} describes how the memory energy barrier height is connected to two important parameters: (a) the parameter update rate; and (b) the learning rate, both of which determine two of the three performance walls, namely, the {\it update-wall} and the {\it consolidation-wall}. For instance, the {\it update-wall} is reflected in the profile of the update-rate $J_n$ for each of the parameters, and Eq. \ref{eq_energyEpsDelta-2} shows how a specific update-profile $J_n$ can be achieved by modulating the barrier-profile in the {\it learning-in-memory} paradigm. Similarly, the learning rate $\epsilon_n$ determines the {\it consolidation-wall}. Several adaptive synaptic models have been proposed~\cite{Fusi2005FusiSynapse,Kirkpatrick2017CatastrophicForgetting} that show how a specific learning-rate profile can lead to optimal information transfer rate between short-term and long-term memories. In the LIM paradigm, the mechanism by which external energy is dissipated to drive the memory/synaptic updates is determined by the gradients provided by the learning algorithm. Like a Lyapunov system, the network or the algorithmic gradient is source of energy and determines the dynamics. {From a thermodynamic computing point of view, the system (or memory) is initialized in a low-entropy state and is guided towards the final state through thermal fluctuations and by energy-barrier modulation.} Thus, like Fig.~\ref{fig_4}(F), {the energy, memory and compute functions are all integrated, a concept originally proposed by Feynmann and Bennett}~\cite{FeynmanCompute1998Feynman,Bennett1982}. {Unlike a traditional computing that requires periodic resets and where entropy production leads to unavoidable energy-loss}~\cite{Wolpert_2020,doi:10.1073/pnas.2321112121}, the LIM approach is a single-shot computation without any resets.  
However, in estimating energy-dissipation for LIM, we have ignored some practical constraints which we point out here. We have assumed that the energy-barrier modulation trajectory is pre-determined which then leads to the desired final state (result of the computation). In practice, it might not be possible to achieve this and would require dissipating additional energy in learning or adapting the modulation procedure. ({Secondly, we have assumed that the energy used to change the energy-barrier height is unrecoverable, resulting in the static term in Eq.~\ref{eq_main_Etotal_general}. In practice, the energy dissipated due to barrier modulation could be recovered but is challenging and would incur additional energy cost.} We have also assumed that the reservoirs driving the memory elements do not dissipate energy which in practice is also difficult to achieve. Larger values of $\gamma$ or a faster update-rate decay implies that a higher $J_{max}$ is needed to satisfy the computational constraint for the training to terminate, i.e., the update-rate $J_n$ should also decay asymptotically to zero. In practice, the maximum update rate $J_{max}$ is determined by the maximum switching frequency ($f_T$) of a physical switching devices. For example, silicon-germanium heterojunction bipolar transistor~\cite{Chakraborty2014SiGeHBT} and an ultrafast optical switch~\cite{Hui2023UltrafastOptics} can exhibit maximum switch-rates close to \SI{1}{\tera\hertz}, equivalently, $J_{max} \approx 10^{12}s^{-1}$.

({From the view point of learning, we assumed in our derivations that the learning gradients are temporally uncorrelated with a constant variance. This is an idealization for obtaining an analytically tractable lower-bound, as described in Eq.\ref{eq_main_weightdecayGD3}. In practice, many commonly used learning mechanisms are known to induce temporal gradient correlations and violate the statistical independence assumptions. Examples of such learning algorithms include momentum and Adam optimizers~\cite{QIAN1999GDMomentum,Kingma2014Adam}, learning-rate schedules and restart schemes~\cite{Loshchilov2017AdamW, Smith20171Cycle}, stateful normalization layers~\cite{ioffe2015batchnormalizationacceleratingdeep}, data-sampling effects~\cite{li2018visualizinglosslandscapeneural} and curvature-induced nonstationarity~\cite{Wu2017TowardsUG}. These correlations could be exploited to reduce the energy dissipation bounds even further, however the analysis would require Monte-Carlo based estimation.}

Irrespective of the energy dissipation mechanisms, LIM energy dissipation lower bounds are based on modulating the memory energy barrier according to Eq.~\ref{eq_energyEpsDelta-2} and Fig.~\ref{fig_3}. Energy barrier modulation supporting the LIM paradigm could be implemented in a variety of physical substrates using emerging memory devices. For instance, recently, we reported a dynamic memory device~\cite{Mehta2022FNsynapse} that could also be used to modulate the memory retention profile and could be an attractive candidate to implement the LIM paradigm. However, note that to approach the fundamental energy limits of training/learning one would need to address all three performance walls. Compute-in-memory (CIM) alternatives where the computation and memory are vertically integrated in massively parallel, distributed architecture offer substantially greater computational bandwidth and energy efficiency in memristive neuromorphic cognitive computing \cite{Wan2022CIMmemristor} approaching the nominal energy efficiency of synaptic transmission in the human brain \cite{Cauwenberghs2013ReverseEngrBrain}.  Resonant adiabatic switching techniques in charge-based CIM~\cite{Karakiewicz2012ResonantArray} further extend the energy efficiency by recycling the energy required to move charge by coupling the capacitive load to an inductive tank at resonance, providing a path towards efficiencies in cognitive computing superior to biology and, in principle, beyond the Landauer limit by overcoming the constraints of irreversible dissipative computing. It is an open question whether the learning-in-memory energy bounded by Eq.~\ref{eq_main_master} could also be at least partially recovered through principles of adiabatic energy recycling.

{However, for practical non-reversible architectures, the dissipation limits are determined by thermodynamic principles. Conventional AI training hardware performance one-to-one mapping between the algorithmic updates and the updates executed on hardware. As a result, the entropy production of the hardware update trajectory starting from the initial state to the final state is practically zero. The energy that is dissipated in the process (by keeping the memory barrier height higher) is to ensure that the entropy does not leak out, i.e., preventing thermal fluctuations from disturbing the logical states in memory consumes energy.} However, this algorithm-to-hardware mapping fails to ignore two general facts about AI training algorithms or optimization algorithms: (a) parameter updates that are guided by the optimization gradients have an inherent error-correcting capability (gradients direct the updates towards the optimal solution), hence paths can absorb fluctuations; and (b) fluctuations in parameter trajectory act as regularization in many AI training algorithms and hence has beneficial effects. So, a more efficient approach would be to leverage stochasticity, i.e., use intrinsic thermal fluctuations rather than avoiding them, as these fluctuations can benefit AI training or optimization algorithms. {The LIM paradigm essentially achieves both by exploiting the combination of thermal fluctuations and memory barrier modulation and in the process dissipating less energy. Studies have been performed that emphasize that stochastic thermodynamics can help analyze such new paradigms that make use of stochasticity rather than opposing it} \cite{doi:10.1073/pnas.2321112121}. 

The use of $\#FLOPS$ as a constraint as opposed to a finite time-of-completion makes the proposed energy-estimation approach agnostic to the specifics of the learning algorithm and the underlying algorithmic efficiency. As a result, the LIM energy dissipation estimates can be extrapolated 
for realistic AI workloads using empirical trends shown in Fig.~\ref{fig_7}. The relationship between model size (number of parameters) and computation complexity (number of FLOPs) was extrapolated from numbers reported in the literature. Prior to a certain model size threshold ($10^9$), the computation grows polynomially with respect to the number of parameters while this trend becomes linear after the development of AI models surpasses the inflection point. While this work did not go in-depth to investigate the quadratic-to-linear phase transition, we can speculate several possible reasons that could be topics of future research. The first reason could be a practical limitation that arises from the model size ($10^9$) at the phase transition. For model size less than $10^9$, the parameters could be directly stored in the main memory and as a result, the energy cost of optimal pair-wise comparison is manageable. Beyond $10^9$ parameters, external storage needs to be accessed, in which case the prohibitive energy cost dictates practical online training algorithms whose complexity grows linearly. The second reasoning for the phase-transition observed in Fig.~\ref{fig_7}A could be more fundamental. It is possible that the quadratic-to-linear transition can be explained using the Tracy-Widom distribution~\cite{majumdar2014tracyWidomUniversality}, which is the universal statistical law underlying phase-transition in complex systems, such as water freezing into ice~\cite{djurivckovic2011water}, graphite transitioning into diamond~\cite{khaliullin2011graphite}, and metals transforming into superconductors~\cite{he2011metal}. In the phase characterized by strong coupling, the system's energy scales quadratically with the number of components/parameters. Conversely, in the phase of weak coupling, the energy is directly related to the count of components/parameters. The training of models also seems to have followed this strong-to-weak trend, deep neural network architectures have become more modular with more distinctive functionality (for example, the multi-head attention block in Large Language Models). 

While the theoretical results described in the paper suggest that the minimum energy required to train a brain-scale AI system using the LIM paradigm is $\sim6$ orders of magnitude lower than the projected energy dissipation for other approaches, the article does not prescribe a specific method to approach this limit. The key assumption that was made in the derivation of LIM lower bound, is mapping of the training problem into a locally convex optimization problem which was then mapped to physical energy through Lyapunov dynamics. Since LIM memory updates are performed through the physical energy gradients that are inherent to the network, the individual LIM units need to be coupled to each other to form a flat memory system. In~\cite{Rahman2023DynamicFNMem} we proposed one such LIM array based on dynamic floating-gate technology, where each memory unit updates itself to minimize the overall energy consumption rather than individual local energy. Also, if the dynamics of specific AI training algorithms can be incorporated into LIM, then the energy dissipation estimates can be lowered even further. For instance, if an $L_2$-norm based regularization is used improve the generalization of AI training, the regularization process is equivalent to a leaky gradient-descent. In~\cite{Rahman2023DynamicFNMem} this synergy was exploited in dynamic floating-gate memories to improve the energy efficiency of training deep neural networks. 



\section*{Data Availability}
The code for simulation and data that support the findings of this article are publicly available at~\cite{Chen2025LIMrepo}.

\begin{acknowledgments}
This work is supported by the National Science Foundation with research grant FET-2208770.
\end{acknowledgments}

\appendix*

\section{}

\subsection{Energy dissipation estimates for a constant energy-barrier (CEB) model}\label{sec_material_convmem}

To estimate a model-agnostic energy cost for memory updates, we assume a linear relationship between the number of training FLOPs ($\#FLOPs$) and the total energy dissipation ($E_{total}$).
This assumption is supported by empirical evidence shown in Fig.~\ref{fig_7}. 
For a given bit-precision ($\#bits$), we get an estimate as
\begin{equation}
    E_{total} \approx \left(\#FLOPs + M \right)\times \#bits\times E_{bit}\label{eq_supp_totalEnergyGeneral}
\end{equation}
where $E_{bit}$ is the energy dissipated per state transition of a single bit. Note that the estimate in Eq.~\ref{eq_supp_totalEnergyGeneral} also takes into account the final energy used for storing $M$ parameters after the training has been completed. 
For a practical memory $E_{bit} = kT \log(1/\delta) \approx \SI{1}{\femto\joule}$ to $\SI{1}{\pico\joule}$. Training a bran-scale AI system with $M = 10^{15}$ and requiring $10^{28}$ FLOPS (according to Fig.~\ref{fig_7}) at a precision of \SI{16}{\bit} would dissipate \qtyrange{e2}{e5}{\tera\joule} of energy. As shown in Fig.~\ref{fig_8}C, the energy dissipation metrics estimated using Eq.~\ref{eq_supp_totalEnergyGeneral} matches the reported energy dissipation for current state-of-the-art AI models within a constant scaling factor.

\subsection{Calibration of LIM hyperparameters}\label{sec_supp_caliberation}
Here we relate the model-specific hyperparameters $\beta$, $\lambda$, and $c$ in Eq.~\ref{eq_main_weightdecayGD5} to an LIM parameter that is model agnostic. For gradient-based learning the parameter $\epsilon_n > 0$ needs to satisfy the following dynamic constraint: 
\begin{eqnarray}
    \lim_{n\to\infty} \epsilon_n &=& 0 \nonumber \\
    \sum_{n=1}^{\infty} \epsilon_n &=& \infty.
    \label{methods_eq_constraints_e}
\end{eqnarray}
While many time-schedules for $\epsilon_n$ satisfying the constaints~\ref{methods_eq_constraints_e} are admissible, an $\epsilon_n = \mathcal{O}\left({1/\sqrt n}\right)$ schedule has been proposed~\cite{Fusi2005FusiSynapse, Benna2016FusiConsolidation,Rahman2023DynamicFNMem} for achieving optimal memory consolidation in benchmark random pattern experiments. 
From Eq.~\ref{eq_main_weightdecayGD3}, substituting $\epsilon_n$ for the optimal schedule,
\begin{equation}
\begin{split}
        \Delta E_n &= \beta kT \sigma_g^2 \sum_{k=1}^n\prod_{m=k+1}^n(1-\lambda\epsilon_m)\epsilon_k^2\\
        &= \beta kT \sigma_g^2 \sum_{k=1}^n\exp\left(\sum_{m=k+1}^n\log(1-\lambda\epsilon_m)\right)\epsilon_k^2
\end{split}
\end{equation}\label{eq_supp_weightdecayGD1}
In most modern ML models, $0<\lambda c\ll1$~\cite{Devlin2018BERT, He2015ResNet}, leading to $\log{\left(1-\frac{\lambda c}{\sqrt m}\right)}\approx-\frac{\lambda c}{\sqrt m}$. Therefore, Eq.~\ref{eq_supp_weightdecayGD1} can be rewritten as
\begin{equation}
\begin{split}
        \Delta E_n &= \beta kT \sigma_g^2 \sum_{k=1}^n\exp\left(-\lambda c(\sqrt{n}-\sqrt{k})\right)\left(\frac{c^2}{k}\right)\\
        &\geq \beta kT \sigma_g^2\frac{c^2}{n}\sum_{k=1}^n\exp\left(-\lambda c(\sqrt{n}-\sqrt{k})\right)\\
\end{split}
\end{equation}\label{eq_supp_weightdecayGD2}
Since we consider the lower bound, we approximate the term $\frac{c^2}{k}$ in the series using its lower bound $\frac{c^2}{n}$. Furthermore, since the summation in Eq.~\ref{eq_supp_weightdecayGD2} is dominated by terms when $k\rightarrow n$, let $p=n-k$, $p\ll n$, we get
\begin{equation}
\begin{split}
        \Delta E_n &\geq \beta kT \sigma_g^2\frac{c^2}{n}\sum_{p=0}^{n-1}\exp\left(-\lambda c\left(\sqrt{n}-\sqrt{n}\sqrt{1-\frac{p}{n}}\right)\right)\\
        &\geq \beta kT \sigma_g^2\frac{c^2}{n}\sum_{k=1}^n\exp\left(-2\lambda c\frac{p}{\sqrt n}\right)\\
        &\geq \beta kT \sigma_g^2\frac{c^2}{n}\frac{1}{1-\exp{\left(-\frac{2\lambda c}{\sqrt n}\right)}}\quad(\text{Geometric series}),
\end{split}
\end{equation}\label{eq_supp_weightdecayGD3}
Given $a_n\rightarrow0$, we can use first order expansion to estimate $\frac{1}{1-\exp{\left(-a_n\right)}}$, leading to
\begin{equation}
\begin{split}
        \Delta E_n 
        &\geq \beta kT \sigma_g^2\frac{c^2}{n}\frac{\sqrt n}{2\lambda c}=C/\sqrt{n},
\end{split}
\end{equation}\label{eq_supp_weightdecayGDFinal}
where $C=\beta kT \sigma_g^2c/{2\lambda}$. For the training to terminate, the update-rate $J_n$ should also decay asymptotically to zero  which leads to the constraint
\begin{equation}
    \lim_{n\to\infty}J_n = 0.\label{methods_eq_constraints_R}
\end{equation}
Again, numerous choices for scheduling $J_n$ to satisfy the constraint~\ref{methods_eq_constraints_R} are possible. Since the update rate $J_n$ proceeds at least as fast as the learning rate $\epsilon_n$, to achieve the asymptotic bound, we make $J_n:=\epsilon_n$. Note that for this condition the total number of updates becomes unbounded or
\begin{equation}
    \sum_{n=1}^{\infty} J_n = \infty.\label{methods_eq_adiabatic_R}
\end{equation}
The asymptotic barrier height $E_{\infty}^0$ can be estimated by inserting the respective schedules for $\epsilon_n$ and $J_n$ in the Eq.~\ref{eq_main_master} which leads to
\begin{equation}
	E_{\infty}^0 \ge \lim_{n \to \infty} kT \log \left[2\sqrt n \cdot\frac{\exp\left(C/\sqrt n\right) - 1}{\exp\left(C/\sqrt n\right) + 1}\right] = kT \log \left(C\right),\label{methods_eq_adiabaticE0} 
\end{equation}
where $C=\beta \sigma^2(g_{d})c/2\lambda$. This asymptotic result is then equated to the modified Landauer's limit~\cite{Bennett1982} which specifies the absolute minimum height of the energy-barrier $E^0 = kT \log(1/\delta)$ that is needed to ensure that a parameter could be stored/updated at a precision $\delta$. 
Thus, from Eq.~\ref{methods_eq_adiabaticE0}, $C = \frac{1}{\delta}$ which makes the bound in Eq.~\ref{eq_main_master} model agnostic as
\begin{equation}
	E_n^0 \ge kT \log \left[\frac{2J_{max}}{J_n}\cdot\frac{\exp\left(\epsilon_n/\delta\right) - 1}{\exp\left(\epsilon_n/\delta\right) + 1}\right].\label{methods_eq_masterEqF}
\end{equation}
The bound now connects the barrier-height $E_n^0$, the update-rate $J_n$, the learning-rate (a memory consolidation parameter) $\epsilon_n$, and the precision of computation/memory retention $\delta$.


\subsection{LIM energy estimates for large AI models}\label{sec_supp_energy_estimates}



For a learning rate schedule $\epsilon_n = 1/\sqrt n$, the update-rate schedule $r_n\sim\frac{J_n}{J_{max}} = 1/n^{1+\gamma}$ with $\gamma >0$ ensures that Eq.~\ref{methods_eq_constraints_R}and~\ref{eq_main_constraints_E0_flops} are satisfied. In LIM$_A$, the energy to drive the memory updates are derived from the algorithmic gradient. Therefore, according to Eq.~\ref{methods_eq_masterEqF} and the asymptotic calibration, $\Delta E_n^{diss} = \epsilon_n \left(kT/\delta\right)$, Eq.~\ref{eq_main_EtotalA} leads to
\begin{eqnarray}
E_{Total} &\approx& \#FLOPs \left[\left(\frac{kT}{\delta}\right) \frac{\sum_{n=1}^{\infty} \epsilon_n r_n}{\sum_{n=1}^\infty r_n}\right]  + M E_{\infty}^0
\nonumber\\
            &=& \#FLOPs \left(\frac{kT}{\delta}\right) \left[\frac{\sum_{n=1}^{\infty} \frac{1}{n^{3/2+\gamma}}}{\sum_{n=1}^{\infty} \frac{1}{n^{1+\gamma}}}\right] + M E_{\infty}^0. \nonumber\\
\label{eq_supp_EtotalA_ex1}
\end{eqnarray}
Eq.~\ref{eq_supp_EtotalA_ex1} can now expressed in an analytic form
\begin{equation}
E_{Total} \approx \#FLOPs \left(\frac{kT}{\delta}\right) \left[\frac{\zeta(3/2+\gamma)}{\zeta(1+\gamma)}\right] + M kT \log \left[ \frac{1}{\delta} \right] \nonumber
\label{eq_supp_EtotalA_ex2}
\end{equation}
where $\zeta(.)$ denotes the Riemann-zeta function.

\subsection{LIM analysis for exponential decay update-rate}\label{sec_supp_exponential_analytical}
In this section, we derive the analytical energy-dissipation estimates for LIM with $\epsilon_n\sim\frac{1}{n}$ and exponentially-decaying update rate schedule $r(n)\sim e^{-\gamma n}$. Similar to what has been shown for a polynomial decay update-rate in the main text,
\begin{equation}
\begin{split}
E_{Total} &\approx M \left(\frac{kT}{\delta}\sum_{n=1}^{\infty} {R_n} \epsilon_n \Delta t + E_{\infty}^0\right) \\
            &= \#FLOPs \left(\frac{kT}{\delta}\right) \left[\frac{\sum_{n=1}^{\infty} \frac{1}{n}e^{-\gamma n}}{\sum_{n=1}^{\infty} e^{-\gamma n}}\right] + M E_{\infty}^0 \\
            &= \#FLOPs \left(\frac{kT}{\delta}\right) \left[(1-e^{\gamma})\log{\left(\frac{e^{\gamma n}}{e^{\gamma}-1}\right)}\right] \\
            &+ M kT \log \left[ \frac{1}{\delta} \right]
\end{split}
\end{equation}\label{supp_eq_main_EtotalA_ex}

\bigskip
\bibliography{main}

\end{document}